\documentclass[runningheads]{llncs}
\usepackage{graphicx}
\usepackage{amsmath,amssymb} % define this before the line numbering.
\usepackage{color}
\usepackage{booktabs}
\usepackage{lib}
\usepackage{wrapfig}
\usepackage{gensymb}
\usepackage{multirow}

\usepackage{floatrow}
\usepackage[width=122mm,left=12mm,paperwidth=146mm,height=193mm,top=12mm,paperheight=217mm]{geometry}
\usepackage{xcolor}
\newfloatcommand{capbtabbox}{table}[][\FBwidth]
\begin{document}
% \linenumbers
\pagestyle{headings}
\mainmatter

\definecolor{Plum}{RGB}{142, 69, 133}
\definecolor{Cyan}{RGB}{0, 255, 255}
\definecolor{Red3}{HTML}{a40000}
\definecolor{Green3}{HTML}{4e9a06}
\newcommand{\todo}[1]{\textcolor{Red3}{TODO: #1}}

\title{
View Synthesis by Appearance Flow
} % Replace with your title

\titlerunning{View Synthesis by Appearance Flow}

\authorrunning{T. Zhou, S. Tulsiani, W. Sun, J. Malik, A. A. Efros}

\author{Tinghui Zhou, Shubham Tulsiani, Weilun Sun,\\ Jitendra Malik, Alexei A. Efros}
\institute{University of California, Berkeley}

\maketitle

\begin{abstract}
%Jitendra's comment on abstract's starting : can try replacing lines 123 by 13'2' instead.

%We address the problem of {\em novel view synthesis}: given an input image, synthesizing new images of the same object or scene observed from arbitrary viewpoints.  Previous approaches either leverage geometric transformations of the original image, or use machine learning to synthesize novel views directly.  In this work, we attempt to combine the two by starting with a learning-based approach but re-formulating it as a pixel copying task that avoids the difficulties of generating pixels from scratch.

%% Attempting big J's version below
We address the problem of {\em novel view synthesis}: given an input image, synthesizing new images of the same object or scene observed from arbitrary viewpoints. We approach this as a learning task but, critically, instead of learning to synthesize pixels from scratch, we learn to \emph{copy} them from the input image.  %%%This allows us to combine previous approaches which either leverage geometric transformations of the original image, or use machine learning to synthesize novel views directly. 
%%In this work, we develop a hybrid approach by but re-formulate it as a pixel copying task that avoids the difficulties of generating pixels from scratch.
%%In this work, we develop a hybrid approach by reformulating the learning problem from pixel synthesis to pixel copying/reuse/transformation?  
Our approach exploits the observation that the visual appearance of different views of the same instance is highly correlated, and such correlation could be explicitly learned by training a convolutional neural network (CNN) to predict \emph{appearance flows} -- 2-D coordinate vectors specifying which pixels in the input view could be used to reconstruct the target view. Furthermore, the proposed framework easily generalizes to multiple input views by learning how to optimally combine single-view predictions. We show that for both objects and scenes, our approach is able to synthesize novel views of higher perceptual quality than previous CNN-based techniques.

%Given one or more images\todo{AE: maybe just say "single image", and then mention more at the end of abstract? TZ: I agree} of an object (or a scene), is it possible to synthesize a new image of the same instance observed from an arbitrary viewpoint? In this paper, we attempt to tackle this problem, known as novel view synthesis,  by re-formulating it as a pixel copying task that avoids the 
%notorious 
%difficulties of generating pixels from scratch. Our approach is built on the observation that the visual appearance of different views of the same instance is highly correlated. Such correlation could be explicitly learned by training a convolutional neural network (CNN) to predict \emph{appearance flows} -- 2-D coordinate vectors specifying which pixels in the input view could be used to reconstruct the target view. We show that for both objects and scenes, our approach is able to generate higher-quality synthesized views with crisp texture and boundaries than previous CNN-based techniques. 
\end{abstract}

\begin{figure}[htb]
    \centering
    \includegraphics[width=\linewidth]{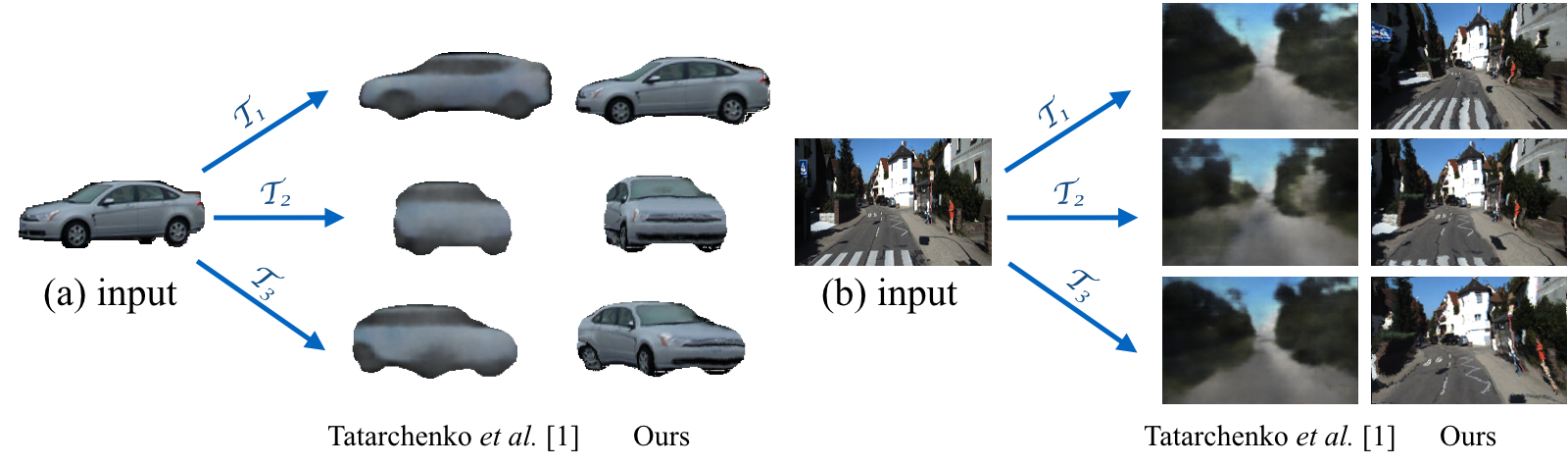}
    \caption{Given an input image, our goal is to synthesize novel views of the same object (left) or scene (right) corresponding to various camera transformations ($T_i$). Our approach, based on learning appearance flows, is able to generate higher-quality results than the previous method that directly outputs pixels in the target view~\cite{tatarchenko2015single}.}
    \figlabel{teaser}
\end{figure}

\section{Introduction}

Consider the car in \figref{teaser}(a). Actually, what you are {\em looking at} is a flat two-dimensional image that is but a projection of the three-dimensional physical car.   Yet, numerous psychophysics experiments tell us that what you are {\em seeing} is not the 2D image but the 3D object that it represents. For example, one classic experiment
demonstrates that people excel at ``mental rotation"~\cite{Shepard701} -- predicting what a given object would look like after a known 3D rotation is applied.  In this paper, we study the computational equivalent of mental rotation called {\em novel view synthesis}.  Given one or more input images of an object or a scene plus the desired viewpoint transformation, the goal is to synthesize a new image capturing this novel view, as shown in \figref{teaser}. 

Besides purely academic interest (how well can this be done?), novel view synthesis has a plethora of practical applications, mostly in computer graphics and virtual reality.    
For example, it could enable photo editing programs like Photoshop to manipulate objects in 3D instead of 2D.  Or it could help create full virtual reality environments based on historic images or video footage.

%Even though this car is just a single, flat, two-dimensional picture, we humans have little trouble thinking of it in 3D and ``mentally rotating"~\cite{mental_rotation} it to imagine how it might look from a different viewing angle.  Even if we can only see one side of the car, our imagination can usually produce a {\em plausible} picture of what its front or back might look like.  

%scientific interest -- can you do view synthesis

%psychological interest -- people seem to do mental rotation. why can human do this task -- recognizing two images depict the same object from different views? Implicit 3D reasoning? Recognition by components?

%graphics interest -- natasha, Tour-into-picture, Derek, VR

The ways that novel view synthesis has been approached in the past fall into two broad categories: geometry-based approaches and learning-based approaches.  Geometric approaches try to first estimate (or fake) the approximate underlying 3D structure of the object, and then apply some transformation to the pixels in the input image to produce the output~\cite{horry1997,oh2001,zhang2002,hoiem2005,zheng2012_cuboid,chen2013_3sweep,kholgade20143d}.  Besides the requirement of somehow estimating the 3D structure, which is a difficult task by itself, the other major downside of these methods is that they produce holes in places where the source image does not have the appropriate visual content (e.g. the back side of an object).  In such cases, various types of texture hole-filling are sometimes used but they are not always effective.  

Learning-based approaches, on the other hand, argue that novel view synthesis is fundamentally a learning problem, because otherwise it is woefully underconstrained. Given a side of a car, there is no way to ever guess what the front of this car looks like, unless the system has observed other fronts of cars so it can make an educated guess.  Such methods typically try, at training time, to build a parametric model of the object class, and then use it at test time, together with the input image, to generate a novel view.   Unfortunately, parametric image generation is an open research topic, and currently the results of such methods are often too blurry (e.g. see~\cite{tatarchenko2015single} in \figref{teaser}). 

In this paper, we propose to combine the benefits of both types of approaches, while also avoiding their pitfalls. Like geometric methods, we propose to use the pixels of the input image as much as possible, instead of trying to synthesize new ones from scratch.  At the same time, we will use a learning-based approach to implicitly capture the approximate geometry of the object, avoiding the explicit estimation of the 3D structure. Our model also learns the appearance correlation between different parts of the object that enables synthesizing the backside of the object. 

Conceptually, our approach is quite simple: we train a deep generative convolutional encoder-decoder model, similar to~\cite{tatarchenko2015single}, but instead of generating RGB values for each pixel in the target view, we generate an {\em appearance flow} vector indicating the corresponding pixel in the input view to steal from.  This way, the model does not need to learn how to generate pixels from scratch -- just where to copy from the input view. In addition to making the learning problem more tractable, it also provides a natural way of preserving the identity and structure of the input instance -- a task typically difficult for conventional learning approaches. We demonstrate the applicability of our approach by synthesizing views corresponding to rotation of objects and egomotion in scenes. We further extend our framework to leverage multiple input views and empirically show the quantitative as well as perceptual improvements obtained with our approach. 

\section{Related work}
\noindent \textbf{Feature learning by disentangling pose and identity.}
Synthesizing novel views of objects can be thought of as decoupling pose and identity and has long been studied as part of feature learning and view-invariant recognition. Hinton \etal~\cite{hinton2011transforming} learned a hierarchy of ``capsules'', computational units that locally transform their input, for generating small rotations to an input stereo pair, and argued for the use of similar units for recognition. More recently, Jaderberg \etal~\cite{jaderberg2015spatial} demonstrated the use of computational layers that perform global spatial transformation over their input features as useful modules for recognition tasks. Jayaraman \etal~\cite{jayaramanIccv2015} studied the task of synthesizing features transformed by ego-motion and demonstrated its utility as an auxiliary task for learning semantically useful feature space. Cheung \etal~\cite{cheung2014discovering} proposed an auto-encoder with decoupled semantic units representing pose, identity \etc and latent units representing other factors of variation and showed that their approach was capable of generating novel views of faces. Kulkarni \etal~\cite{kulkarni2015deep} introduced a similarly motivated variational approach for decoupling and manipulating the factors of variation for images of faces. While the feature-learning approaches convincingly demonstrated the ability to disentangle factors of variation, the view manipulations demonstrated were typically restricted to small rotations or categories with limited shape variance like digits and faces.

\noindent \textbf{CNNs for view synthesis.}
 A recent interest in learning to synthesize views for more challenging objects under diverse view variations has been driven by the ability of Convolutional Neural Networks (CNNs)~\cite{neocognitron,Lecun1989} to function as image decoders. Dosovitiskiy \etal~\cite{DosovitskiyChairs} learned a CNN capable of functioning as a renderer: given an input graphics code containing identity, pose, lighting \etc their model could render the corresponding image of a chair. Yang \etal~\cite{yang2015weakly} and Tatarchenko \etal~\cite{tatarchenko2015single} built on this work using the insight that the graphics code, instead of being presented explicitly, can be implicitly captured by an example source image along with the desired transformation. Yang \etal~\cite{yang2015weakly} learned a decoder to obtain implicit pose and identity units from the input source image, applied the desired transformation to the pose units, and used a decoder CNN to render the desired view. Concurrently, Tatarchenko \etal~\cite{tatarchenko2015single} followed a similar approach without the explicit decoupling of identity and pose to obtain similar results. A common module in these approaches is the use of a decoder CNN to generate the pixels corresponding to the transformed view from an implicit/explicit graphics code. Our work demonstrates that predicting appearance flows instead of pixels leads to significant improvements.

\noindent \textbf{Geometric view synthesis.}
An alternative paradigm for synthesizing novel views of an object is to explicitly model the underlying 3D geometry.  In cases when more than one input view is available, modern multi-view stereo algorithms (see Furukawa and Hernandez~\cite{mvs_tutorial} for an excellent tutorial) have demonstrated results of impressive visual quality. However, these methods fundamentally rely on finding visual correspondences -- pixels that is in common across the views -- so they break down when there are only a couple of views from very different viewpoints. In cases when only a single view is available, user interaction had typically been needed to help define a coarse geometry for the object or scene~\cite{horry1997,oh2001,zhang2002,zheng2012_cuboid,chen2013_3sweep}.  More recently, large Internet collections of stock 3D shape models have been leveraged to get 3D geometry for a wide range of common objects. For example, Kholgade \etal~\cite{kholgade20143d} obtained realistic renderings of novel views of an object by transferring texture from the corresponding 3D model, though they required manual annotation of the exact 3D model and its placement in the image. Rematas \etal~\cite{RematasArxiv2016} employed a similar technique after automatically inferring the closest 3D model from a shape collection as well as explicitly obtaining pose via a learnt system to situate the 3d model in the image. Their approach, however, is restricted to rendering the closest model in the shape collection instead of the original object. Su \etal~\cite{swyg-3afsnvo-15} overcome this restriction by interpolating between several similar models from the shape collections, though they only demonstrate their technique for generating HOG~\cite{dalal2005histograms} features for novel views. Unlike the CNN based learning approaches, these geometry-based methods require access to a shape collection during inference and are limited by the intermediate bottlenecks of inferring pose and retrieving similar models.

\noindent \textbf{Image-based Rendering.}
%\noindent \textbf{Palette-based Image Generation.}
The idea of directly re-using the pixels from available images to generate new views has been popular in computer graphics. Debevec \etal~\cite{debevec1996modeling} used the underlying geometry to composite multiple views for rendering novel views. Lightfield/lumigraph~\cite{levoy1996light,gortler1996lumigraph} rendering presented an alternate setup where a structured, dense set of views is available. Buehler \etal~\cite{buehler2001unstructured} presented a unifying framework for these image-based rendering techniques. The recent DeepStereo work by Flynn \etal~\cite{flynn2015deepstereo} is a learning-based extension that performs compositing through learned geometric reasoning using a CNN, and can generate intermediate views of a scene by interpolating from a set of surrounding views. While these methods yield high-quality novel views, they do so by composting the corresponding input image rays for each output pixel and can therefore only generate already seen content, (\eg they cannot create the rear-view of a car from available frontal and side-view images).

\noindent \textbf{Texture Synthesis and Epitomes.}
Reusing pixels of the input image to synthesize new visual context is also at the heart of non-parametric texture synthesis approaches.  In texture synthesis~\cite{efros1999texture,barnes2009patchmatch}, the synthesized image is pieced together by combining samples of the input texture image in a visually consistent way, whereas for texture transfer~\cite{hertzmann2001image,efros2001image}, an additional constraint aims to make the overall result also mimic a secondary ``source'' image.  
A related line of work uses \textit{epitomes}~\cite{jojic2003epitomic} as a generative model for a set of images. The key idea is to use a condensed image as a palette for sampling patches to generate new images. In a similar spirit, our approach can be thought of as generating novel views of an object using the original image as an epitome.

%Others. Building view graph... Render for CNN... Pulkit ICCV

\section{Approach}
Our approach to novel view synthesis is based on the observation that the appearance (texture, shape, color, etc.) of different views of the same object/scene is highly correlated, and in many cases even a single input view contains rich amount of information for inferring various novel views. For instance, given the side view of a car, one could extract appearance properties such as the 3D shape, body color, window layout and wheel types of the query instance that are sufficient for reconstructing many other views.  

In this work, we \emph{explicitly} infer the appearance correlation between different views of a given object/scene by training a convolutional neural network that takes 1) an input view and 2) a desired viewpoint transformation, and predicts a dense \emph{appearance flow field} (AFF) that specifies how to reconstruct the target view using pixels from the input view. Specifically, for each pixel $i$ in the target view, the appearance flow vector $f^{(i)} \in \mathbb{R}^2$ specifies the coordinate at the input view where the pixel value is sampled to reconstruct pixel $i$. The notion of appearance flow field is closely related to the nearest neighbor field (NNF) in PatchMatch~\cite{barnes2009patchmatch}, except that NNF is explicitly defined on a distance function between two patches, while our appearance flow field is the output of a CNN after end-to-end training for cross-view reconstruction.

The benefits of predicting the appearance flow field over raw pixels of the target view are three-fold: 1) It alleviates the perceptual blurriness in images generated by CNN trained with $L_p$ loss. By constraining the CNN to only utilize pixels available in the input view, we are able to avoid the undesirable local minimum attained by predicting the mean (when $p=2$) colors around texture/edge boundaries that lead to blurriness in the resulting image (e.g. see \secref{exp} for empirical comparison). 2) The color identity of the instance is preserved by construction since the synthesized view is reconstructed using only pixels from the same instance; 3) The appearance flow field enables intuitive interpretation of the network output since we can visualize exactly how the target view is constructed with the input pixels (e.g. see \figref{flows}).

We first describe our training objective and the network architecture for the setting of a single input view in \secref{obj}, and then present a simple extension in \secref{multi_view} that allows the network to learn how to combine individual predictions when multiple input views are available.

\subsection{Learning view synthesis via appearance flow}
\seclabel{obj}
Recall that our goal is to train a CNN that, given an input view $I_s$ and a relative viewpoint transformation $T$, synthesizes the target view $I_t$ by sampling pixels from $I_s$ according to the predicted appearance flow field. This can be formalized as minimizing the following objective:
\begin{equation}
\text{minimize}~\hspace{0.05in}\sum_{<I_s,I_t,T> \in \mathcal{D}} \|I_t - g(I_s,T)\|_p,\hspace{0.1in} \text{subject to}\hspace{0.1in} g^{(i)} (I_s,T) \in \{I_s \}, \forall i~,
\label{eq:obj}
\end{equation}
where $\mathcal{D}$ is the set of training tuples, $g(\cdot)$ refers to the CNN whose weights we wish to optimize, $\|\cdot\|_p$ denotes the $L_p$ norm\footnote{We use $p=1$ in all our experiments, but similar results can be obtained with $L_2$ norm as well.}, and $i$ indexes over pixels of the synthesized view. Internally, the CNN computes a dense flow field $f$, where each element $f^{(i)} \triangleq (x^{(i)}, y^{(i)})$ specifies the pixel sampling location (in the coordinate frame of the input view) for constructing the output $g^{(i)}(I_s, T)$. To allow end-to-end training via stochastic gradient descent when $f^{(i)}$ falls into a sub-pixel coordinate, we rewrite the constraint of Eq.~\ref{eq:obj} in the form of bilinear interpolation:
\begin{equation}
    g^{(i)}(I_s, T) = \sum_{q \in \{\text{neighbors of } (x^{(i)}, y^{(i)})\}} I_s^{(q)} (1 - |x^{(i)} - x^{(q)}|) (1 - |y^{(i)} - y^{(q)}|)~,
\end{equation}
where $q$ denotes the 4-pixel neighbors (top-left, top-right, bottom-left, bottom-right) of $(x^{(i)}, y^{(i)})$. This is also known as differentiable image sampling with a bilinear kernel, and its (sub)-gradient with respect to the CNN parameters could be efficiently computed~\cite{jaderberg2015spatial}.

\noindent\textbf{Network architecture}~~~~Our view synthesis network~(\figref{single_net}) follows a similar high-level design as~\cite{yang2015weakly} and~\cite{tatarchenko2015single} with three major components: 
\begin{enumerate}
    \item Input view encoder -- extracts relevant features (e.g. color, pose, texture, shape, etc.) of the query instance (6 conv + 2 fc layers).
    \item Viewpoint transformation encoder -- maps the specified relative viewpoint to a higher-dimensional hidden representation (2 fc layers).
    \item Synthesis decoder -- assembles features from the two feature encoders, and outputs the appearance flow field that reconstructs the target view with pixels from the input view (2 fc + 6 uconv layers).
\end{enumerate}
 All the convolution, fully-connected and fractionally-strided/up-sampling convolution (uconv) layers are followed by rectified linear units except for the last flow decoder layer. 

\noindent\textbf{Foreground prediction}~~~~For synthesizing object views, we also train another network that predicts the foreground segmentation mask of the target view. The architecture is the same as the synthesis network in~\figref{single_net}, except that in this case the last layer predicts a per-pixel binary classification mask ($0$ is background and $1$ is foreground), and the network is trained with cross-entropy loss. At test time, we further apply the predicted foreground mask to the synthesized view. 

\begin{figure}[t!]
    \centering
    \includegraphics[width=\linewidth]{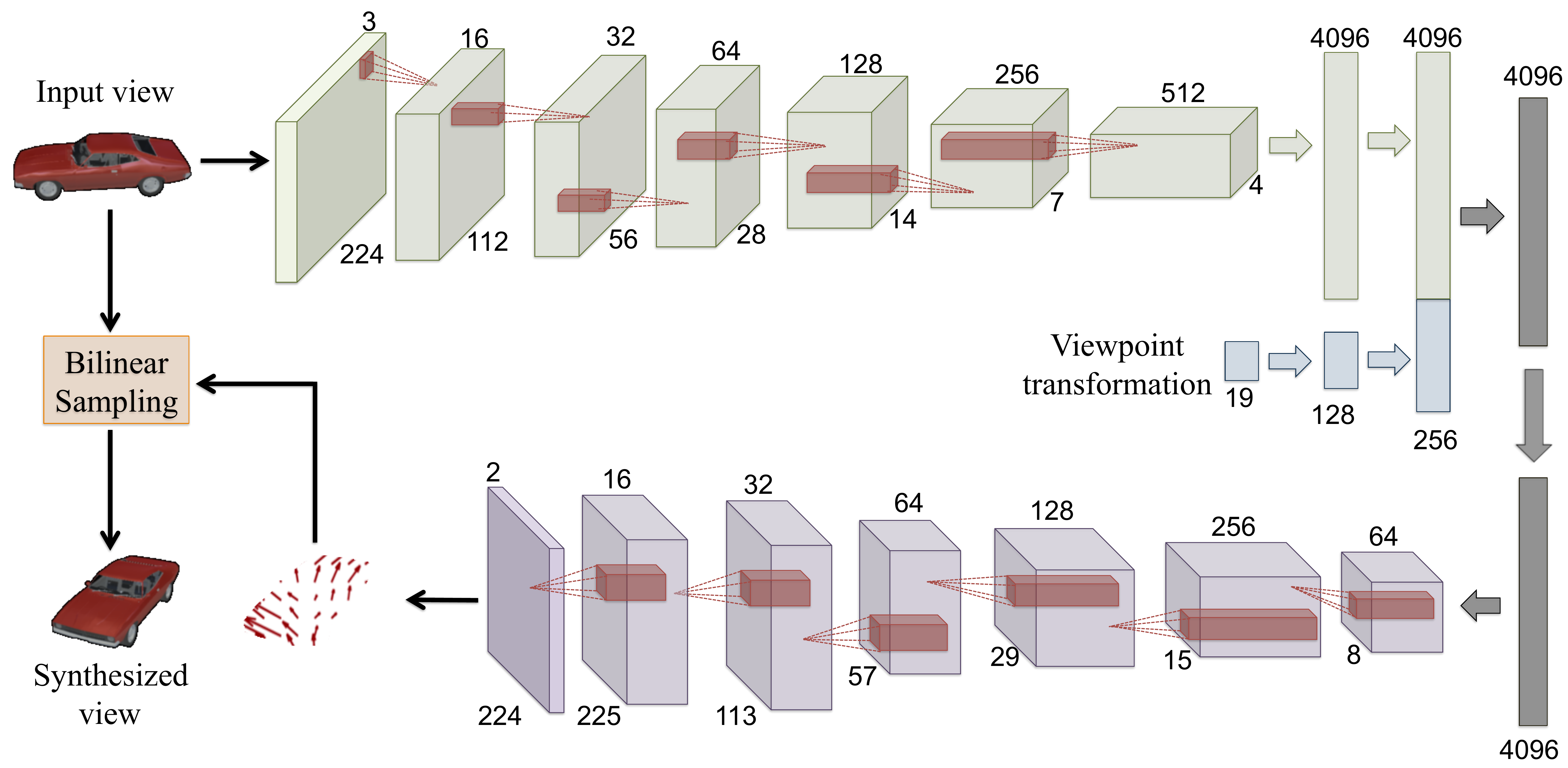}
    \caption{Overview of our single-view network architecture. We follow an encoder-decoder framework where the input view and the desired viewpoint transformation are first encoded via several convolution and fully-connected layers, and then a decoder consisting of two fully-connected and six up-sampling convolution layers outputs an appearance flow field, which in conjunction with the input view yields the synthesized view through a bilinear sampling layer. All the layer weights are learned end-to-end through back-propagation.}
    \figlabel{single_net}
\end{figure}

\subsection{Learning to leverage multiple input views}
\seclabel{multi_view}
A single view of the object sometimes might not contain sufficient information for inferring an arbitrary target view. For instance, it would be very challenging to infer the texture details of the wheel spoke given only the frontal view of a car, and similarly, the side view of a car contains little to none information about the appearance of the head lights. Thus, it would be ideal to develop a mechanism that could leverage the individual strength of different input views to synthesize target views that might not be feasible with any input view alone. 

To achieve this, we modify our view synthesis network to also output a \emph{soft} confidence mask $C_j$ that indicates per-pixel prediction quality using input view $s_j$, which could be implemented by adding an extra output channel to the last decoder layer. The confidence masks for all input views are further normalized to sum to one at each pixel location: $\bar{C}_j^{(i)} = C_j^{(i)}/\sum_{k=1}^NC_k^{(i)}$, where $N$ denotes the number of input views. Intuitively, $\bar{C}_j^{(i)}$ is an estimator of \emph{relative} prediction quality using input view $j$ at pixel $i$, and by using $\bar{C}_j$ as a hypothesis selection mask, the final joint prediction is simply a weighted combination of hypotheses predicted by different input views: $\sum_{j=1}^N \bar{C}_j * g(I_{s_j}, r_j)$. \figref{multi_net} illustrates the architecture of our multi-view network that is also end-to-end learnable. 

\noindent\textbf{Comparison with DeepStereo~\cite{flynn2015deepstereo}}~~~~While the general idea of learning hypothesis selection for view synthesis has been recently explored in~\cite{flynn2015deepstereo}, there are a few key differences between our framework and~\cite{flynn2015deepstereo}: 1) We do not require projecting the input image stack onto a planesweep volume that prohibits their method from synthesizing pixels that are invisible in the input views (i.e. view extrapolation); 2) Unlike~\cite{flynn2015deepstereo}, who have a fixed number of input views, our multi-view network is more flexible at both training and test time as it could take in an \emph{arbitrary} number of input views for joint prediction, which is particularly beneficial when the number of input views varies at test time.

\begin{figure}[t!]
    \centering
    \includegraphics[width=\linewidth]{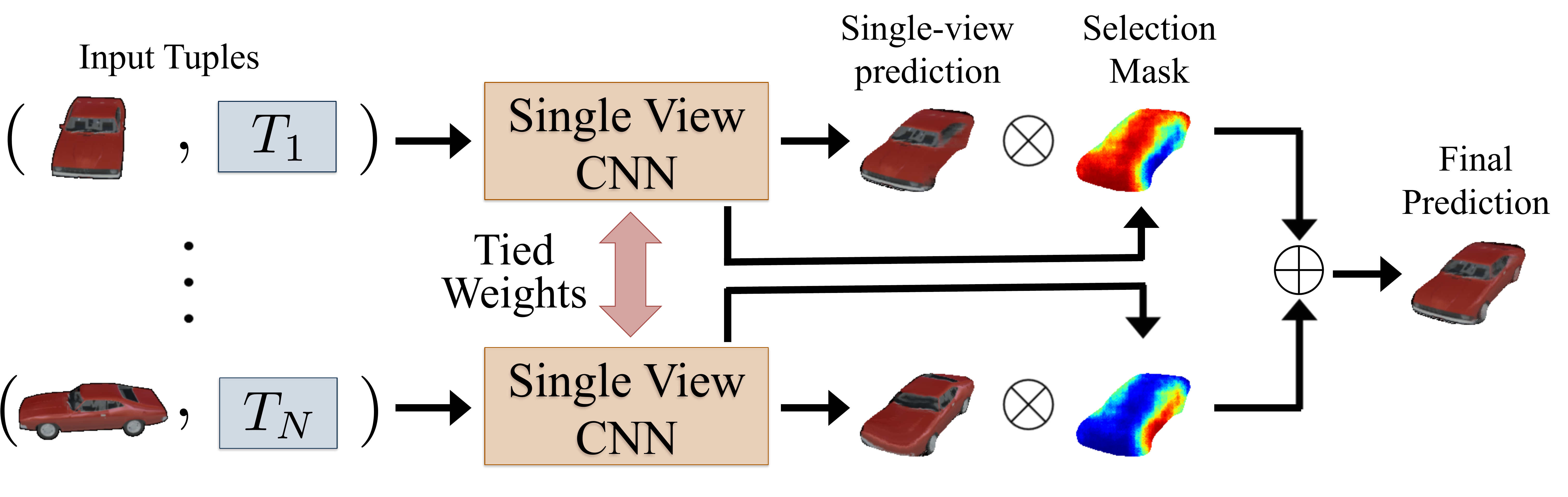}
    \caption{Overview of our multi-view network architecture ($\otimes$: per-pixel product, $\oplus$: per-pixel normalized sum). For each input view, we use a single-view CNN (same as \figref{single_net} but with an extra output channel) with shared weights to independently predict the target view as well as a per-pixel selection/confidence mask. The final target view prediction is obtained by linearly combining the predictions from each view weighted by the selection masks.}
    \figlabel{multi_net}
\end{figure}

\section{Experiments}
\seclabel{exp}
To evaluate the performance of our view synthesis approach, we conduct experiments on both objects (\emph{car},  \emph{chair} and \emph{aeroplane}) and urban city scenes (KITTI~\cite{Geiger2012CVPR}). Our main baseline is the recent work of Tatarchenko et al~\cite{tatarchenko2015single} that synthesizes novel views by training a CNN to directly generate pixels. For fair comparison, we use the same number of network layers for their method and ours, and for experiments on multiple input views we extend their method to output hypothesis selection masks as described in \secref{multi_view}.

\noindent\textbf{Network training details}~~~~We train the networks using a modified version of Caffe~\cite{jia2014caffe} to support the bilinear sampling layer. We use the ADAM solver~\cite{kingma2014adam} with $\beta_1=0.9, \beta_2=0.999,$ initial learning rate of $0.0001$, step size of $50,000$ and a step multiplier $\gamma=0.5$.

\subsection{Novel view synthesis for objects}

\noindent\textbf{Data setup}~~~~We train and evaluate our view synthesis CNN for objects using the ShapeNet database~\cite{shapenet2015}. In particular, we split the available shapes ($7,497$ cars and $700$ chairs\footnote{The original ShapeNet core release contains a total of $6,778$ chair models. However, a majority of the models are of low visual quality (e.g. texture-less), and we only keep a subset of $700$ high-quality ones for our experiments.}) of each category into $80\%$ for training and $20\%$ for testing. For each shape, we render a total of $504$ viewing angles (azimuth ranges from $0\degree$ to $355\degree$, and elevation ranges from $0\degree$ to $30\degree$, both at steps of $5\degree$) with fixed camera distance. For simplicity, we limit the viewpoint transformation for CNN to a discrete set of $19$ azimuth variations ranging from $-180\degree$ to $+180\degree$ at steps of $20\degree$, and encode the transformation as a $19$-D one-hot vector. 

At each training iteration, we randomly sample a batch of $<I_s, I_t, T>$ tuples from the training split for the single-view setting, and $<I_{s_1}, I_{s_2}, I_t, T_1, T_2>$ tuples for the multi-view setting, where $T_i$ denotes the relative viewpoint transformation between $I_{s_i}$ and $I_t$, and $T_i$ is randomly sampled from the set of valid transformations. For each category, we construct a test set of $20,000$ tuples by following the same sampling procedure above, except that the shapes are now sampled from the test split.

\noindent\textbf{Appearance flows versus direct pixel generation}~~~~Our first experiment compares the view synthesis performance of our appearance flow approach with the direct pixel generation method by~\cite{tatarchenko2015single} under the single input view setting. 

\figref{synth_pair} compares the view synthesis results using different methods on examples from the test set of two categories (\emph{car} and \emph{chair}). Overall, our prediction tends to be much sharper and matches the ground-truth better than the baseline. In particular, our synthesized views using appearance flows are able to maintain detailed textures and edge boundaries that are lost in direct pixel generation despite both networks are trained with the same loss function.

\begin{figure}[t!]
\centering
\includegraphics[width=\linewidth]{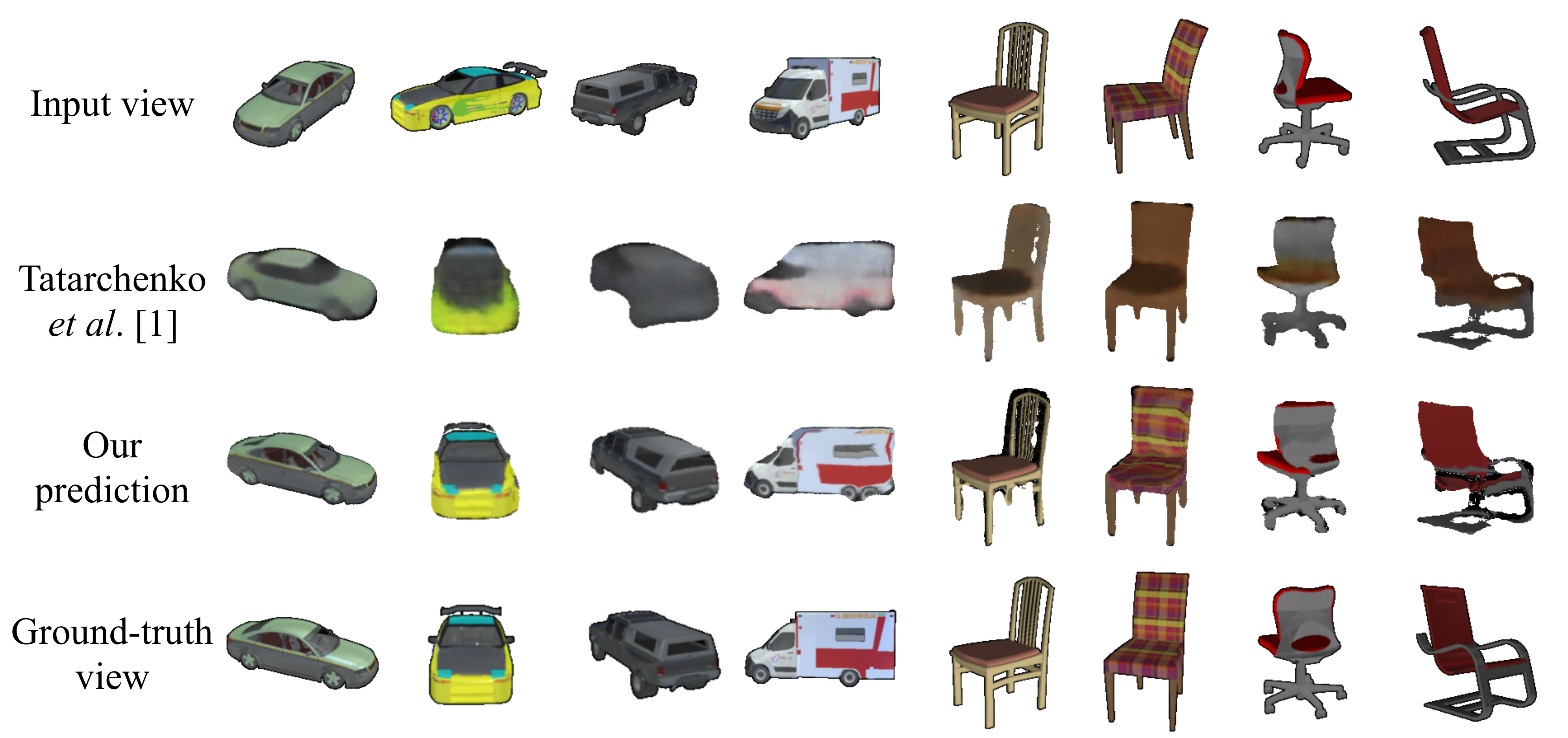}
\caption{Comparison of our single-view synthesis results with the baseline method~\cite{tatarchenko2015single} on cars (left) and chairs (right). Our prediction tends to be consistently better at preserving high-frequency details (e.g. texture and edge boundaries) than the baseline.}
\figlabel{synth_pair}
\end{figure}

For quantitative evaluation we measure the mean pixel $L_1$ error between the predicted views and the ground-truth on the foreground regions. As shown in~\tableref{numbers}, our method outperforms the baseline in both categories (\emph{car} and \emph{chair}). We further analyze the error statistics by computing the pairwise cross-view confusion matrix for both methods, which measures how predictive/informative a given view is for synthesizing another view (see the visualization in~\figref{confusion}). The error statistics suggest that our method is especially strong in synthesizing views that share significant number of common pixels with the input view (within $\pm 45\degree$ azimuth variation from the input view -- the diagonals in the plot) or along the corresponding symmetry planes (off-diagonals) that typically exhibit high appearance correlation with the input view (e.g. synthesizing the right view from the left view of a car), and slightly weaker than direct pixel generation in views that do not share much in common (e.g. from frontal to the side or rear views). 

% \begin{wraptable}{r}{5.5cm}
\begin{table}[t!]
\centering
\begin{tabular}{ccccc}
\toprule
 Input & Method & Car & Chair & KITTI\tabularnewline
\midrule
\multirow{2}{*}{Single-view} & Tatarchenko~\etal~\cite{tatarchenko2015single} & $0.404$ & $0.345$ & $0.492$ \tabularnewline
  & Ours & $\mathbf{0.368}$  & $\mathbf{0.323}$ & $\mathbf{0.471}$ \tabularnewline
\midrule
\multirow{2}{*}{Multi-view} & Tatarchenko~\etal~\cite{tatarchenko2015single} & $0.385$ & $0.334$ & $0.471$ \tabularnewline
  & Ours & $\mathbf{0.285}$ & $\mathbf{0.248}$ & $\mathbf{0.409}$ \tabularnewline
\bottomrule
\end{tabular}
\caption{Mean pixel $L_1$ error between the ground-truth and predictions by different methods. Lower is better.}
\tablelabel{numbers}
\figlabel{wrap-tab:1}
\end{table}
% \end{wraptable} 

\begin{figure}[t!]
    \centering
    \includegraphics[width=\linewidth]{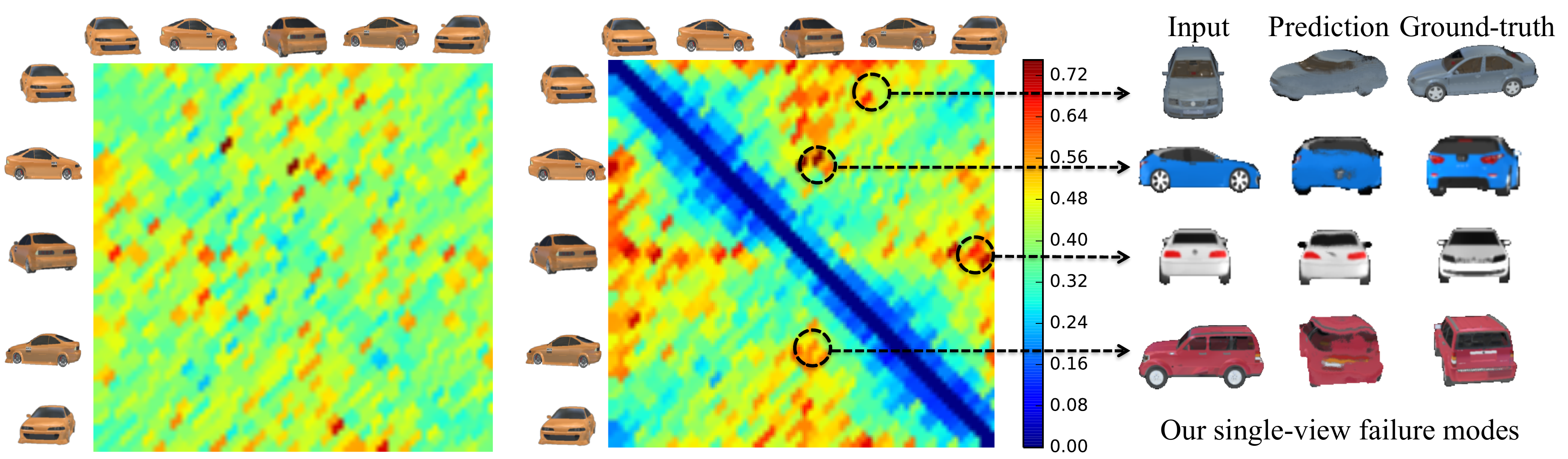}
    \caption{Visualization of error statistics for generating novel views from a single input view on the \emph{car} category. The heatmaps ({\color{blue} blue}--low, {\color{red} red}--high) depict the mean pixel error for obtaining the target view (columns) from the input view (rows) for the baseline~\cite{tatarchenko2015single} (left) and our approach(middle). Some common failure modes of our method are visualized on the right.}
    \figlabel{confusion}
\end{figure}

Interestingly though, when we conduct perceptual studies comparing the visual similarity between predicted views and the ground-truth, our method is far ahead of the baseline across the entire spectrum of the cross-view predictions. More specifically, we randomly sampled $1,000$ test tuples, and asked users on Amazon Mechanical Turk to select the prediction that looks more similar to the ground-truth. We average the responses over $5$ unique turkers for each test tuple, and find that $95\%$ of the time our prediction is chosen over the baseline for cars and $93\%$ for chairs, suggesting that the $L_1$ metric might not fully reflect the strength of our method. 

One additional benefit of predicting appearance flows is that it allows intuitive visualization and understanding of exactly how the synthesized view is constructed. For instance, \figref{flows} shows sample appearance flow vectors predicted by our method. It is interesting to note that the appearance flows do not necessarily correspond to anatomically/symmetrically corresponding parts. For example, while the top-right pixels of the first car in \figref{flows} transfer appearance from their corresponding location in the source image, the pixels in the back wheel are generated using the front wheel of the source image.

\begin{figure}[t!]
\centering
\includegraphics[width=\linewidth]{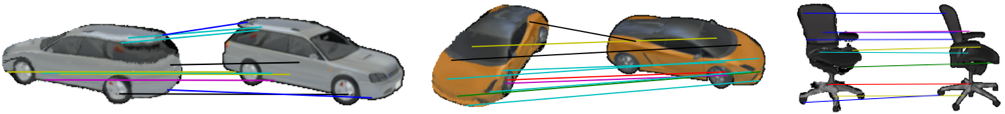}
\caption{Sample appearance flow vectors predicted by our method. For randomly sampled points in the generated target image (left), the lines depict the corresponding appearance flow to the source image (right).}
\figlabel{flows}
\end{figure}

\noindent\textbf{Multi-view versus single-view}~~~~In this experiment, we evaluate the synthesis performance of using multiple input views (two in this case). It turns out that having multiple input views is much more beneficial for our approach than for the baseline, as our synthesis error drops significantly compared to the single-view setting while less so for the baseline (see \tableref{numbers}). This indicates that predicting appearance flows allows more effective utilization of different prediction hypotheses. \figref{multi_view} shows sample visualization of how our multi-view synthesis network automatically combines high-quality predictions from individual input views to construct the final prediction. 

\begin{figure}[t!]
    \centering
    \includegraphics[width=\linewidth]{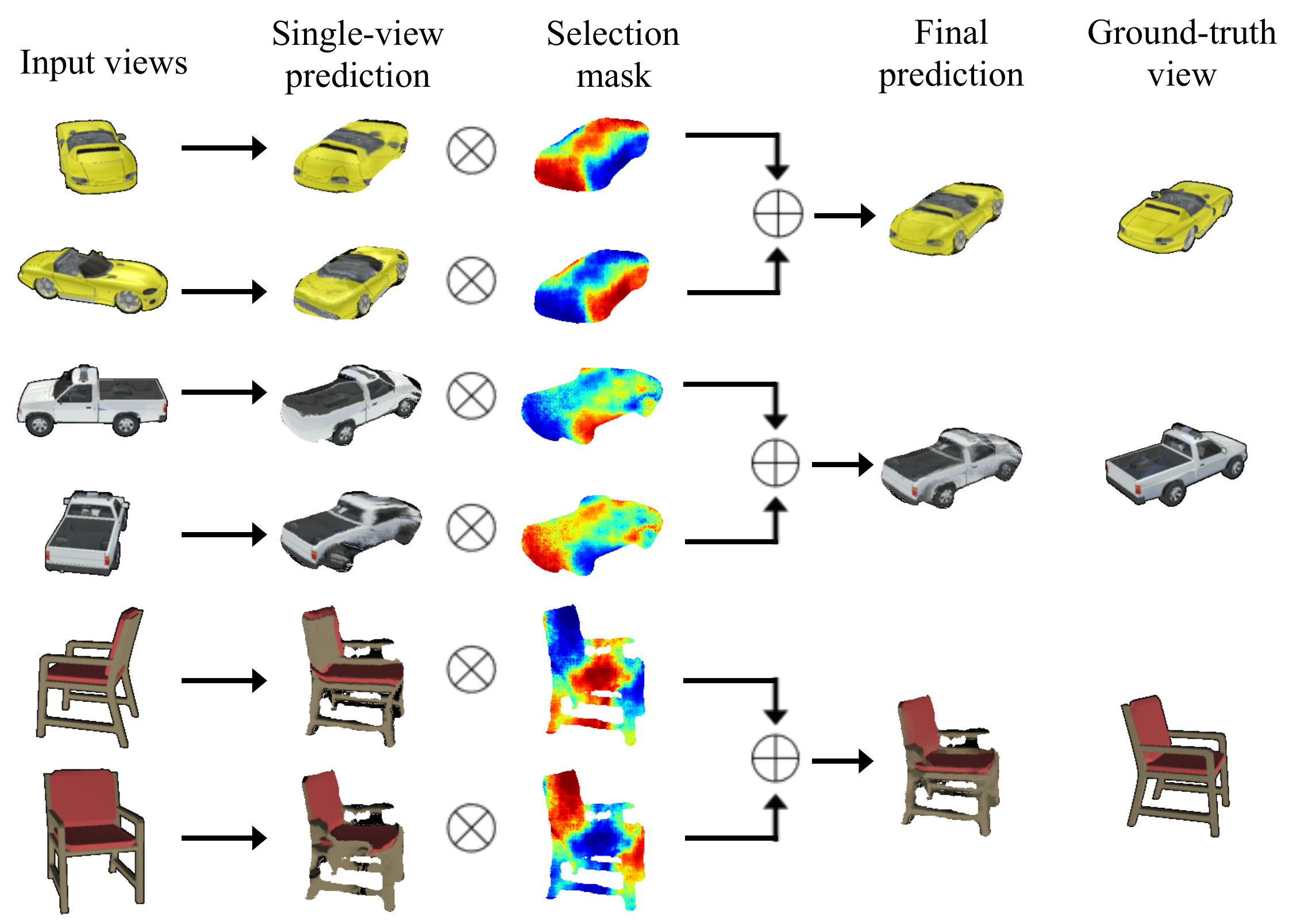}
    \caption{View synthesis examples using our multi-view network. Each input view makes independent prediction of a candidate target view as well as a selection/confidence mask ({\color{blue} blue}--low, {\color{red} red}--high). The final prediction is obtained by linearly combining the single-view predictions with weights normalized across the selection masks. Typically, the final prediction is more similar to the ground-truth than any independent prediction.}
    \figlabel{multi_view}
\end{figure}

\noindent\textbf{Results on PASCAL VOC~\cite{pascal-voc-2012} images}~~~~Although our synthesis network is trained on rendered synthetic images, it also exhibits potentials in generalizing to real images. In order to use our learnt models for synthesizing views for objects in PASCAL VOC, we require some pre-processing to ensure input statistics similar to the rendered training set. We therefore re-scale the input image to have similar number of foreground pixels as objects in the training set with the same aspect ratio.
We visualize and compare a few example synthesis results on segmented PASCAL VOC images in \figref{pascal}. 

\begin{figure}[t!]
    \centering
    \includegraphics[width=\linewidth]{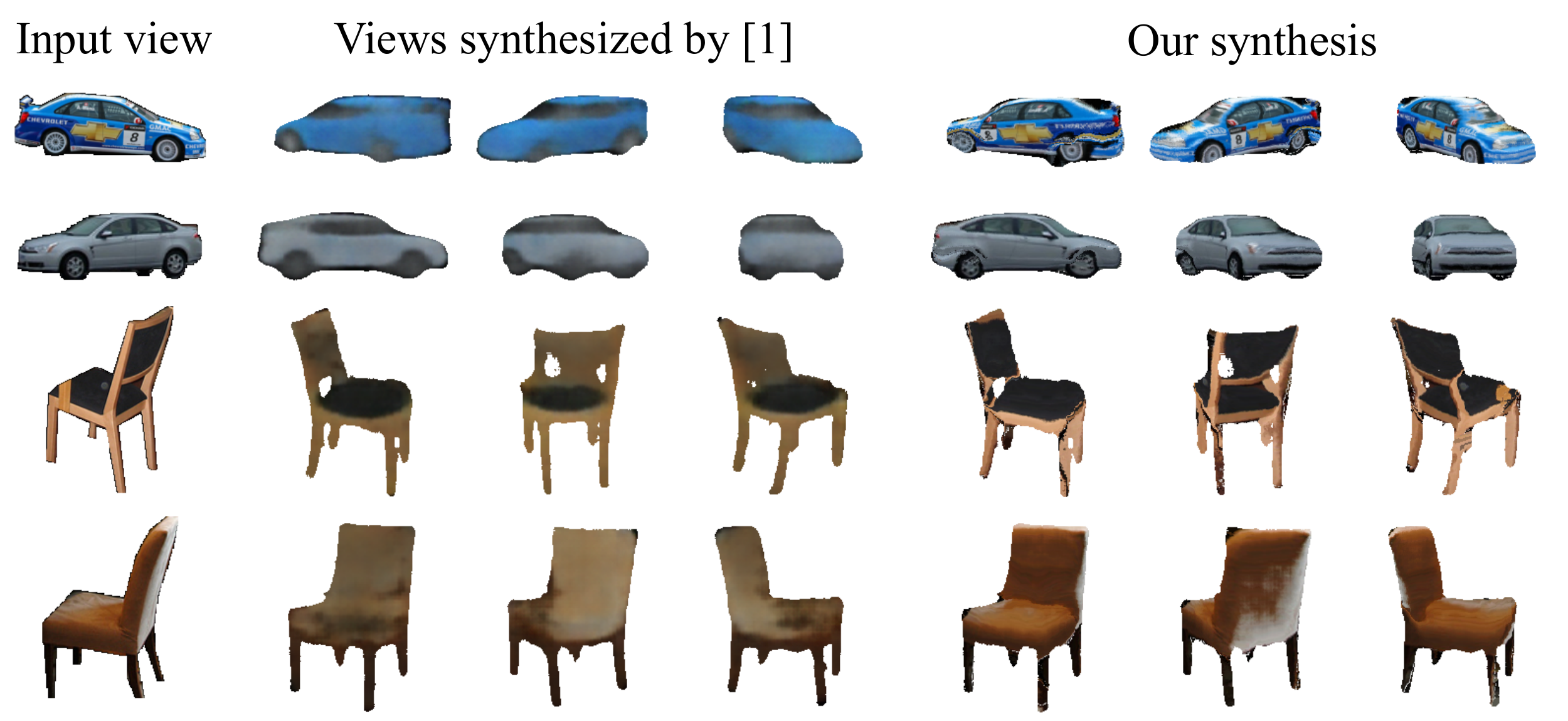}
    \caption{View synthesis results for segmented objects in the PASCAL VOC dataset. Our method generalizes better and yields more realistic results than the baseline~\cite{tatarchenko2015single}.}
    \figlabel{pascal}
\end{figure}

\subsection{Novel view synthesis for scenes}

\noindent\textbf{Data setup}~~~~We evaluate our view synthesis CNN for scenes using the KITTI dataset~\cite{Geiger2012CVPR}, which provides odometry and image sequences taken during $11$ short trips of a car travelling through urban city scenes. We split the $11$ sequences into $9$ for training and $2$ for testing. The viewpoint transformation is computed using the odometry data by taking the difference between the $3\times 4$ transformation matrices (Z-axis pointing forward) of the input and target frames, resulting in a $12$-D vector of continuous values. 

To sample a tuple for the single-view setting, we first randomly sample a sequence ID and then a input frame and a target frame within the sequence that are separated by at most $\pm 10$ frames. For the multi-view setting, we sample an additional input view that is also at most $\pm 10$ frames away from the target view. We randomly sample $10$ tuples for training at each iteration and $20,000$ tuples for testing following the above procedure.
\\
\noindent\textbf{Comparison with direct pixel generation}~~~~Similar to the evaluation on objects, we measure the mean pixel $L_1$ error between the predicted views and the ground-truth. As shown in \tableref{numbers}, our method significantly outperforms the baseline~\cite{tatarchenko2015single} on both single-view and multi-view settings. The advantage is also visualized in~\figref{kitti}, where we compare the predictions made by both methods on the single-view setting. Overall, our prediction tends to preserve the texture details and edge boundaries of objects depicted in the scene (Row $1$--$3$), but sometimes might lead to severe distortions on failure cases (e.g. the last row).

\begin{figure}[t!]
    \centering
    \includegraphics[width=\linewidth]{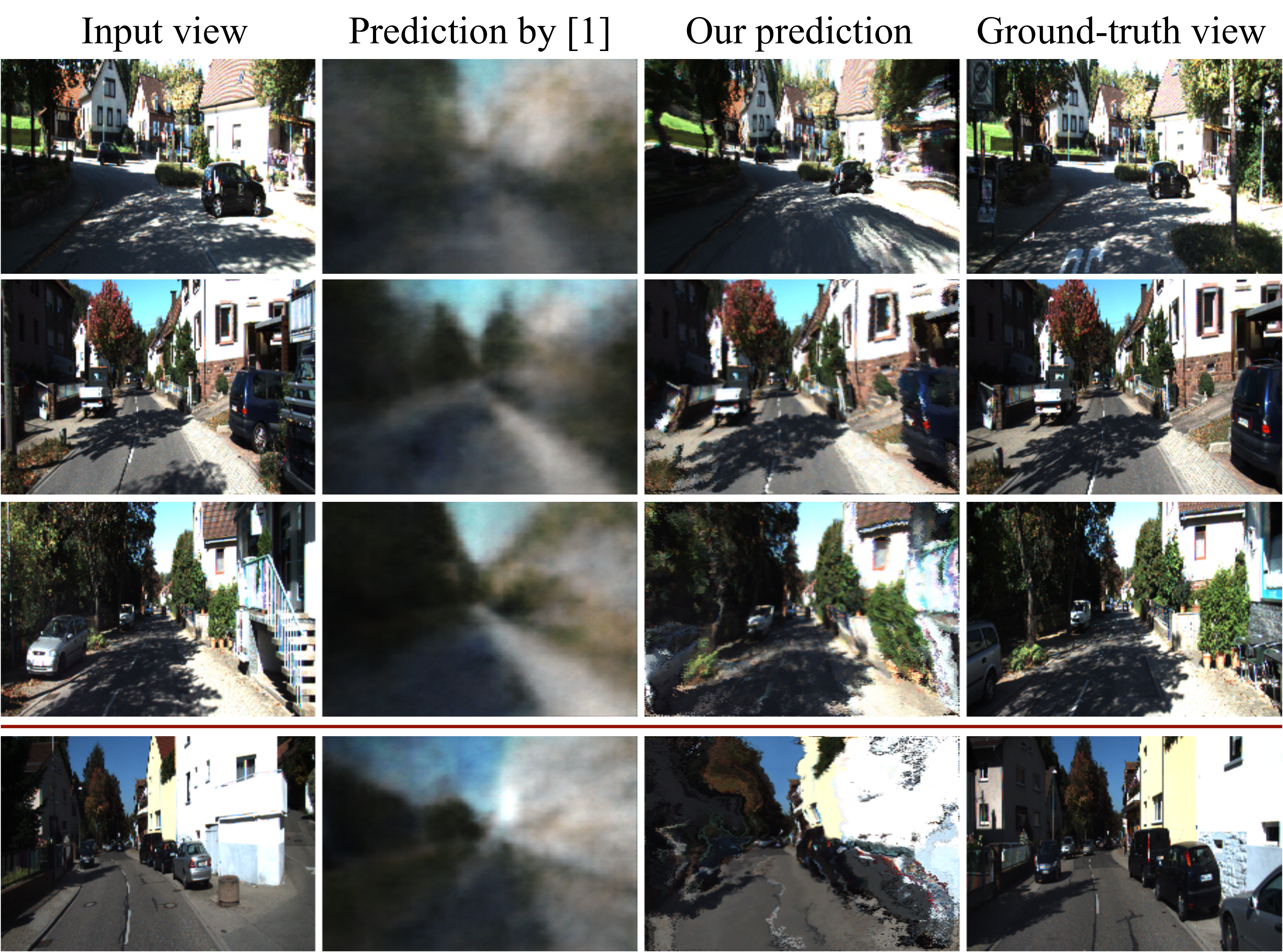}
    \caption{View synthesis results on the KITTI dataset~\cite{Geiger2012CVPR}. Our method typically preserves the scene structure and details of the objects in the synthesized view better than the baseline (a failure case is shown in the last row).}
    \figlabel{kitti}
\end{figure}

\section{Discussion}

%\todo{potential extensions and applications: multi-scale loss, video frame prediction, inpainting, etc... Could synthesize high-res images by upsampling flows}
%\todo{Could synthesize high-res images by upsampling flows}
We have presented a framework that re-parametrizes image synthesis as predicting the appearance flow field between the input image(s) and the output, and demonstrated its successful application to novel view synthesis.
%We have presented a framework that re-parametrizes pixel generation by predicting the appearance flow-field between the input image(s) and the output
%and demonstrated its application for synthesizing novel views of an object (or a scene). 
%
%We have presented a new framework for learning to synthesize novel views of an object (or a scene) by predicting the appearance flow field between the input view and the target view.Despite good performance on various benchmark evaluations, our method is by no means close to solving the problem of view synthesis in the general case. A number of major challenges are yet to be addressed:
%Concerning the task of view synthesis,
But despite good performance on various benchmark evaluations, our method is by no means close to solving the problem in the general case. A number of major challenges are yet to be addressed:
\begin{itemize}
    \item Our current method is incapable of hallucinating pixel values not present in the input view.  While this is not as bad is it sounds (since the color palette of a typical image is quite rich), it would be beneficial to develop a mechanism that combines the hallucination capability of pixel generation CNN and the detail-preserving property of our flow-based synthesis.
    \item Empirically we observe that our network sometimes struggles in learning long-range appearance correlations, since the gradients derived from the flows are quite local. We conducted preliminary experiments with multi-scale reconstruction loss, and found it to alleviate the gradient locality to some extent.
    % \item Our object view synthesis network relies on rendered views of synthetic shapes for training. But simulating complex real-world environments (including illumination, background, occlusion, etc.) is still an open research problem. One way to address this would be via explicit modeling of the domain shift in order to fully generalize to real images.
    \item While the academic community around view synthesis is growing rapidly, we are still missing large-scale datasets of diverse real-world objects/scenes and a proper metric ($L_1$ pixel error is certainly not ideal) for measuring research progress.
    \item All the existing learning-based view synthesis approaches assume knowing the category of the object. An interesting direction is to develop a method that is category-agnostic, and once learned, can be applied to any real-world image. 
\end{itemize}

Finally, we believe that our technique of leveraging appearance flows is also applicable to tasks beyond novel view synthesis, including image inpainting, video frame prediction, modeling effect of actions, super-resolution, \etc

\section*{Acknowledgements}
We thank Philipp Kr\"ahenb\"uhl and Abhishek Kar for helpful discussions. 
This work was supported in
part by NSF award IIS-1212798, Intel/NSF Visual and Experiential Computing award IIS-1539099 and a Berkeley Fellowship. We gratefully acknowledge NVIDIA corporation for the donation of GPUs used for this research.

\bibliographystyle{splncs}
\bibliography{egbib}

\begin{thebibliography}{10}

\bibitem{tatarchenko2015single}
Tatarchenko, M., Dosovitskiy, A., Brox, T.:
\newblock Single-view to multi-view: Reconstructing unseen views with a
  convolutional network.
\newblock arXiv preprint arXiv:1511.06702 (2015)

\bibitem{Shepard701}
Shepard, R.N., Metzler, J.:
\newblock Mental rotation of three-dimensional objects.
\newblock Science (1971)

\bibitem{horry1997}
Horry, Y., Anjyo, K.I., Arai, K.:
\newblock Tour into the picture: using a spidery mesh interface to make
  animation from a single image.
\newblock In: Proceedings of the 24th annual conference on Computer graphics
  and interactive techniques. (1997)

\bibitem{oh2001}
Oh, B.M., Chen, M., Dorsey, J., Durand, F.:
\newblock Image-based modeling and photo editing.
\newblock In: Proceedings of the 28th annual conference on Computer graphics
  and interactive techniques. (2001)

\bibitem{zhang2002}
Zhang, L., Dugas-Phocion, G., Samson, J.S., Seitz, S.M.:
\newblock Single-view modelling of free-form scenes.
\newblock The Journal of Visualization and Computer Animation (2002)

\bibitem{hoiem2005}
Hoiem, D., Efros, A.A., Hebert, M.:
\newblock Automatic photo pop-up.
\newblock ACM transactions on graphics (TOG) (2005)

\bibitem{zheng2012_cuboid}
Zheng, Y., Chen, X., Cheng, M.M., Zhou, K., Hu, S.M., Mitra, N.J.:
\newblock Interactive images: cuboid proxies for smart image manipulation.
\newblock ACM Transactions on Graphics (TOG) (2012)

\bibitem{chen2013_3sweep}
Chen, T., Zhu, Z., Shamir, A., Hu, S.M., Cohen-Or, D.:
\newblock 3-sweep: Extracting editable objects from a single photo.
\newblock ACM Transactions on Graphics (TOG) (2013)

\bibitem{kholgade20143d}
Kholgade, N., Simon, T., Efros, A.A., Sheikh, Y.:
\newblock 3d object manipulation in a single photograph using stock 3d models.
\newblock ACM Transactions on Graphics (TOG) (2014)

\bibitem{hinton2011transforming}
Hinton, G.E., Krizhevsky, A., Wang, S.D.:
\newblock Transforming auto-encoders.
\newblock In: Artificial Neural Networks and Machine Learning--ICANN.
\newblock (2011)

\bibitem{jaderberg2015spatial}
Jaderberg, M., Simonyan, K., Zisserman, A.,  et~al.:
\newblock Spatial transformer networks.
\newblock In: Advances in Neural Information Processing Systems. (2015)

\bibitem{jayaramanIccv2015}
Jayaraman, D., Grauman, K.:
\newblock {Learning image representations tied to egomotion}.
\newblock In: IEEE International Conference on Computer Vision. (2015)

\bibitem{cheung2014discovering}
Cheung, B., Livezey, J.A., Bansal, A.K., Olshausen, B.A.:
\newblock Discovering hidden factors of variation in deep networks.
\newblock arXiv preprint arXiv:1412.6583 (2014)

\bibitem{kulkarni2015deep}
Kulkarni, T.D., Whitney, W.F., Kohli, P., Tenenbaum, J.:
\newblock Deep convolutional inverse graphics network.
\newblock In: Advances in Neural Information Processing Systems. (2015)

\bibitem{neocognitron}
Fukushima, K.:
\newblock {N}eocognitron: {A} self-organizing neural network model for a
  mechanism of pattern recognition unaffected by shift in position.
\newblock Biological Cybernetics (1980)

\bibitem{Lecun1989}
LeCun, Y., Boser, B., Denker, J., Henderson, D., Howard, R.E., Hubbard, W.,
  Jackel, L.D.:
\newblock Backpropagation applied to hand-written zip code recognition.
\newblock In: Neural Computation. (1989)

\bibitem{DosovitskiyChairs}
A.Dosovitskiy, J.T.Springenberg, T.Brox:
\newblock Learning to generate chairs with convolutional neural networks.
\newblock In: IEEE International Conference on Computer Vision and Pattern
  Recognition. (2015)

\bibitem{yang2015weakly}
Yang, J., Reed, S.E., Yang, M.H., Lee, H.:
\newblock Weakly-supervised disentangling with recurrent transformations for 3d
  view synthesis.
\newblock In: Advances in Neural Information Processing Systems. (2015)

\bibitem{mvs_tutorial}
Furukawa, Y., Hern{\'a}ndez, C.:
\newblock Multi-view stereo: A tutorial.
\newblock Foundations and Trends{\textregistered} in Computer Graphics and
  Vision \textbf{9} (2015)

\bibitem{RematasArxiv2016}
Rematas, K., Nguyen, C., Ritschel, T., Fritz, M., Tuytelaars, T.:
\newblock Novel views of objects from a single image.
\newblock arXiv preprint arXiv:1602.00328 (2015)

\bibitem{swyg-3afsnvo-15}
Su, H., Wang, F., Yi, L., Guibas, L.:
\newblock 3d-assisted image feature synthesis for novel views of an object.
\newblock In: International Conference on Computer Vision. (2015)

\bibitem{dalal2005histograms}
Dalal, N., Triggs, B.:
\newblock Histograms of oriented gradients for human detection.
\newblock In: IEEE Conference on Computer Vision and Pattern Recognition.
  (2005)

\bibitem{debevec1996modeling}
Debevec, P.E., Taylor, C.J., Malik, J.:
\newblock Modeling and rendering architecture from photographs: A hybrid
  geometry-and image-based approach.
\newblock In: Proceedings of the 23rd annual conference on Computer graphics
  and interactive techniques. (1996)

\bibitem{levoy1996light}
Levoy, M., Hanrahan, P.:
\newblock Light field rendering.
\newblock In: Proceedings of the 23rd annual conference on Computer graphics
  and interactive techniques, ACM (1996)  31--42

\bibitem{gortler1996lumigraph}
Gortler, S.J., Grzeszczuk, R., Szeliski, R., Cohen, M.F.:
\newblock The lumigraph.
\newblock In: Proceedings of the 23rd annual conference on Computer graphics
  and interactive techniques, ACM (1996)  43--54

\bibitem{buehler2001unstructured}
Buehler, C., Bosse, M., McMillan, L., Gortler, S., Cohen, M.:
\newblock Unstructured lumigraph rendering.
\newblock In: Proceedings of the 28th annual conference on Computer graphics
  and interactive techniques, ACM (2001)  425--432

\bibitem{flynn2015deepstereo}
Flynn, J., Neulander, I., Philbin, J., Snavely, N.:
\newblock Deepstereo: Learning to predict new views from the world's imagery.
\newblock In: IEEE Conference on Computer Vision and Pattern Recognition.
  (2016)

\bibitem{efros1999texture}
Efros, A.A., Leung, T.K.:
\newblock Texture synthesis by non-parametric sampling.
\newblock In: Computer Vision, 1999. The Proceedings of the Seventh IEEE
  International Conference on. Volume~2., IEEE (1999)  1033--1038

\bibitem{barnes2009patchmatch}
Barnes, C., Shechtman, E., Finkelstein, A., Goldman, D.:
\newblock Patchmatch: A randomized correspondence algorithm for structural
  image editing.
\newblock ACM Transactions on Graphics (TOG) (2009)

\bibitem{hertzmann2001image}
Hertzmann, A., Jacobs, C.E., Oliver, N., Curless, B., Salesin, D.H.:
\newblock Image analogies.
\newblock In: Proceedings of the 28th annual conference on Computer graphics
  and interactive techniques, ACM (2001)  327--340

\bibitem{efros2001image}
Efros, A.A., Freeman, W.T.:
\newblock Image quilting for texture synthesis and transfer.
\newblock In: Proceedings of the 28th annual conference on Computer graphics
  and interactive techniques, ACM (2001)  341--346

\bibitem{jojic2003epitomic}
Jojic, N., Frey, B.J., Kannan, A.:
\newblock Epitomic analysis of appearance and shape.
\newblock In: IEEE International Conference on Computer Vision. (2003)

\bibitem{Geiger2012CVPR}
Geiger, A., Lenz, P., Urtasun, R.:
\newblock Are we ready for autonomous driving? the kitti vision benchmark
  suite.
\newblock In: IEEE Conference on Computer Vision and Pattern Recognition.
  (2012)

\bibitem{jia2014caffe}
Jia, Y., Shelhamer, E., Donahue, J., Karayev, S., Long, J., Girshick, R.,
  Guadarrama, S., Darrell, T.:
\newblock Caffe: Convolutional architecture for fast feature embedding.
\newblock arXiv preprint arXiv:1408.5093 (2014)

\bibitem{kingma2014adam}
Kingma, D., Ba, J.:
\newblock Adam: A method for stochastic optimization.
\newblock arXiv preprint arXiv:1412.6980 (2014)

\bibitem{shapenet2015}
Chang, A.X., Funkhouser, T., Guibas, L., Hanrahan, P., Huang, Q., Li, Z.,
  Savarese, S., Savva, M., Song, S., Su, H., Xiao, J., Yi, L., Yu, F.:
\newblock {ShapeNet: An Information-Rich 3D Model Repository}.
\newblock Technical Report arXiv:1512.03012 [cs.GR], Stanford University ---
  Princeton University --- Toyota Technological Institute at Chicago (2015)

\bibitem{pascal-voc-2012}
Everingham, M., Van~Gool, L., Williams, C.K.I., Winn, J., Zisserman, A.:
\newblock The {PASCAL} {V}isual {O}bject {C}lasses {C}hallenge 2012 {(VOC2012)}
  {R}esults.
\newblock
  http://www.pascal-network.org/challenges/VOC/voc2012/workshop/index.html

\end{thebibliography}
\end{document}